\documentclass[runningheads]{llncs}
\usepackage{graphicx}

\usepackage{xspace}
\usepackage{tikz}
\usepackage{comment}
\usepackage{amsmath,amssymb} 
\usepackage{color}
\usepackage{booktabs}
\usepackage{multirow}
\newcommand{\tabincell}[2]{\begin{tabular}{@{}#1@{}}#2\end{tabular}}

\def\ie{\emph{i.e.}}

\usepackage[accsupp]{axessibility}  

\usepackage[breaklinks=true,letterpaper=true,colorlinks,bookmarks=false]{hyperref}

\begin{document}
\pagestyle{headings}
\mainmatter
\def\ECCVSubNumber{3647}  

\title{Towards 3D Scene Understanding by Referring Synthetic Models} 


\titlerunning{Towards 3D Scene Understanding by Referring Synthetic Models}
%
\author{Runnan Chen\inst{1} \and
Xinge Zhu\inst{2} \and
Nenglun Chen\inst{1} \and
Dawei Wang\inst{1} \and
Wei Li\inst{3} \and
Yuexin Ma\inst{4} \and
Ruigang Yang\inst{3} \and
Wenping Wang\inst{1,5}
}
\authorrunning{Chen et al.}
%
\institute{The University of Hong Kong \and The Chinese University of Hong Kong \and
Inceptio \and
ShanghaiTech University \and
Texas A\&M University}
\maketitle

\vspace*{-3ex}
\begin{abstract}
Promising performance has been achieved for visual perception on the point cloud. However, the current methods typically rely on labour-extensive annotations on the scene scans. In this paper, we explore how synthetic models alleviate the real scene annotation burden, \ie , taking the labelled 3D synthetic models as reference for supervision, the neural network aims to recognize specific categories of objects on a real scene scan (without scene annotation for supervision). The problem studies how to transfer knowledge from synthetic 3D models to real 3D scenes and is named Referring Transfer Learning (RTL). The main challenge is solving the model-to-scene (from a single model to the scene) and synthetic-to-real (from synthetic model to real scene's object) gap between the synthetic model and the real scene. To this end, we propose a simple yet effective framework to perform two alignment operations. First, physical data alignment aims to make the synthetic models cover the diversity of the scene's objects with data processing techniques. Then a novel \textbf{convex-hull regularized feature alignment} introduces learnable prototypes to project the point features of both synthetic models and real scenes to a unified feature space, which alleviates the domain gap. These operations ease the model-to-scene and synthetic-to-real difficulty for a network to recognize the target objects on a real unseen scene. Experiments show that our method achieves the average mAP of 46.08\% and 55.49\% on the ScanNet and S3DIS datasets by learning the synthetic models from the ModelNet dataset. Code will be publicly available.
\vspace*{-2ex}
\end{abstract}

\begin{figure*}
  \vspace*{-1.5ex}
  \centerline{\includegraphics[width=1\textwidth]{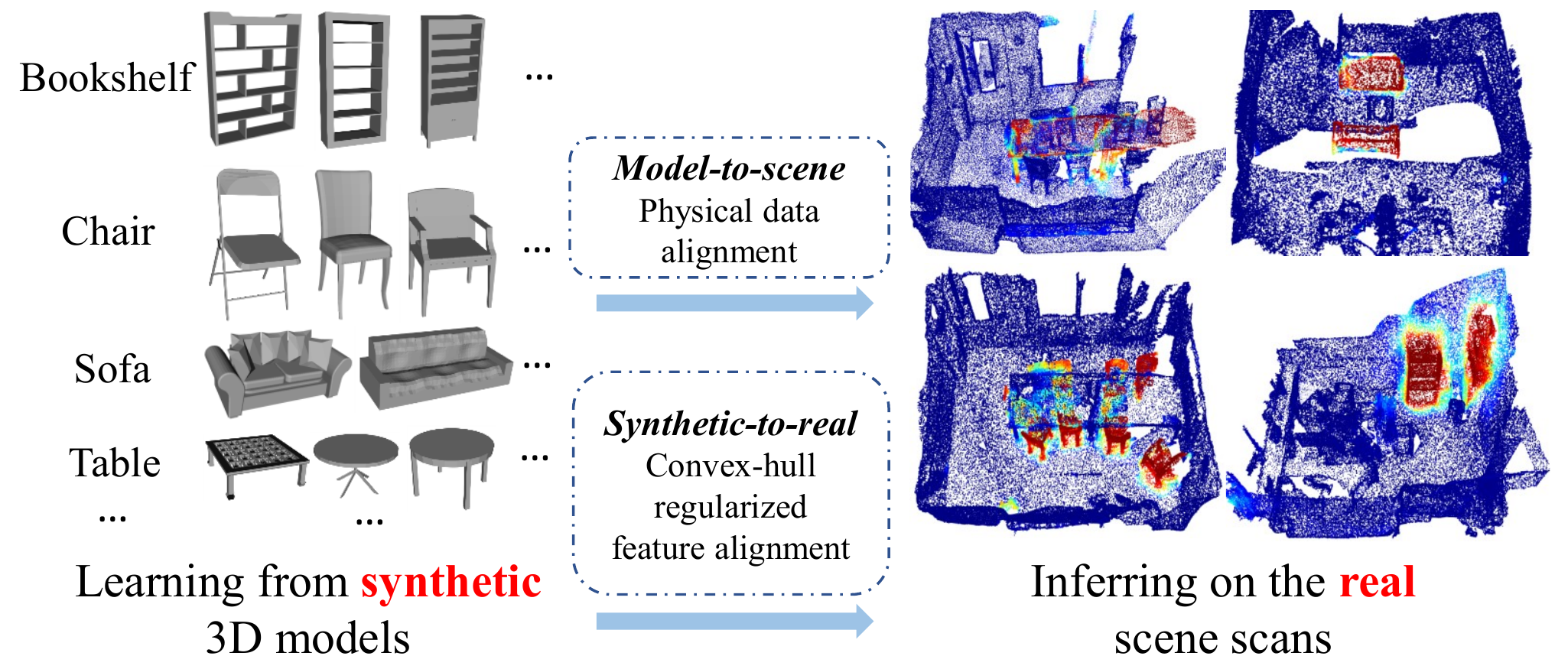}}
  \vspace*{-3ex}
  \caption{We study the problem of transferring knowledge from synthetic 3D models to real 3D scenes, which is named Referring Transfer Learning (RTL). The network aims to recognize specific categories of objects in a real unseen scene by only learning from the synthetic models (without labelled scenes for supervision). RTL emphasizes solving the gaps of model-to-scene (from a single model to the scene) and synthetic-to-real (from synthetic model to real scene's object) between the synthetic model and the real scene. To this end, we propose two alignment operations (physical data alignment and convex-hull regularized feature alignment) to handle the above issues.}
  \label{fig:teaser}
  \vspace{-5ex}
\end{figure*}

\vspace*{-3ex}
\section{Introduction}
Visual perception on the point cloud is fundamental for 3D scene understanding. However, most current works are scene- and domain-specific, indicating that a network trained on a particular dataset works well in the corresponding scenarios with a similar distribution. Thus, the network requires massive data labelling to handle different scenarios and domains. Based on this consideration, this paper investigates the problem of the neural network's model-to-scene and synthetic-to-real generalization ability to reduce the annotation burden for 3D scene understanding, \ie, given the labelled synthetic models as reference, our goal is to recognize specific categories of objects on a real scene. Note that there is no manually labelled scene for supervision. We name it Referring Transfer Learning (RTL) as it refers to synthetic models for supervision and transfers knowledge from synthetic models to real scenes. The model-to-scene means that the network learns from individual 3D models and infers on different scene scans. synthetic-to-real indicates the domain gap between synthetic 3D models and real scenes' objects. The challenges are shown in Fig.~\ref{fig:teaser}. The purpose is to study how to perform effective and efficient model-to-scene and synthetic-to-real knowledge transferring from synthetic models to the real 3D world.

RTL is apart from the existing problem setting. First, it differs naturally from supervised semantic segmentation \cite{ouaknine2021multi,chen2021hierarchical,zhu2021cylindrical,li2021lidar}, which requires labour-extensive scene annotation and is often limited to very few scenarios. A similar setting is the unsupervised domain adaptation \cite{saenko2010adapting,wang2020alleviating,tzeng2017adversarial,qin2019pointdan}, which transfers the knowledge from the source domain to the target domain where target labels are not necessary. Compared with RTL, we also study the model-to-scene generalization that transfers 3D models to crowded scenes. Lastly, unlike traditional optimization-based methods that leverages synthetic models for object detection \cite{song2014sliding}, replacement \cite{litany2017asist} and reconstruction \cite{li2015database,nan2012search}. Our RTL emphasizes the neural network's model-to-scene and synthetic-to-real generalization ability, which is under-explored.

In this paper, we instantiate RTL as a point-wise classification problem (semantic segmentation). The main challenges are the model-to-scene and the synthetic-to-real gap. Specifically, unlike the synthetic models are independent and complete, the objects in a scene are varied in pose, size and location, nearby other objects, and often partially observed. Besides, they also vary in geometric properties, where the objects in a real scan are irregular and noisy due to the limitation of scanning equipment while the synthetic models are clean. As the first learning-based method investigates this problem to the best of our knowledge, our framework is simple yet effective to handle the gaps. It mainly includes two alignment operations, \ie, physical data alignment and convex-hull regularized feature alignment. 

Specifically, inspired by recent 3D data augmentation studies \cite{li2020pointaugment,chen2020pointmixup,zhang2017mixup,zhang2021pointcutmix,nekrasov2021mix3d}, we design a series of data processing approaches to perform the physical data alignment on synthetic models to match the real-world scene, including rotation, scaling, cropping, and mix up with other models and the scene. Besides, there is a domain gap between the synthetic models and the real scene's objects, which indicates that they (the same semantic synthetic and real models) may not share a common feature space. This motivated us to project their feature into a unified space for alleviating the domain gap. Inspired by the convex hull theory~\cite{rockafellar2015convex}, \ie, any convex combinations of the points must be restrained in the convex hull, we propose a novel module to project point features into the convex hull space that regularizes their distributions. The basic idea is to set a group of learnable prototypes as the support points to formulate a convex hull, a closure and compact feature space. The prototypes are designed to learn the base structural elements of 3D models (Fig. \ref{fig:visual_sturcture} (B)), which are used as 'intermedium' to project all point features into the convex hull feature space. In this context, the projected point feature (for both synthetic and real models) is the combination (a convex-hull combination) of the shared base structural elements, restrained in a unified feature space, which alleviate the domain gap. In this way, the physical data alignment and the convex-hull regularized feature alignment ease the model-to-scene and synthetic-to-real difficulty, causing a better generalization ability to recognize the target objects when inferring a real scene.

We conduct the experiments on ModelNet \cite{wu20153d}, ScanNet \cite{dai2017scannet}, and S3DIS datasets \cite{2017arXiv170201105A}, where ModelNet provides the synthetic models for supervision, and the scene scans in ScanNet and S3DIS are for evaluation. Our method achieves the average mAP of 46.08\% and 55.49\% on the ScanNet and S3DIS datasets without manually annotated scene scans for supervision. Furthermore, it can be a potential pretext task to improve the performance of the downstream tasks for 3D scene understanding.

The contributions of our work are as follows.
\vspace*{-1ex}
\begin{itemize}
\item {We formulate and investigate the problem of transferring knowledge from the synthetic 3D models to real 3D scenes, called Referring Transfer Learning (RTL). And we propose a simple yet efficient framework to solve model-to-scene and synthetic-to-real issues.}
\item {We propose a novel convex-hull regularization to handle the domain gap by projecting all point features into a unified feature space.}
\item {Our method achieves promising results to infer the objects of target categories in real scenes on two datasets.}
\end{itemize}
\section{Related Work}
\paragraph{\textbf{Deep Learning on Point Cloud}}
Point cloud, as a 3D data representation, has been used extensively in various 3D related tasks, such as shape classification \cite{liu2021pointguard}, 3D segmentation \cite{ouaknine2021multi,chen2021hierarchical,zhu2021cylindrical,cui2021tsegnet,hong2021lidar}, 3D detection \cite{li2021lidar,zhang2020h3dnet,qi2019deep,zhu2020ssn,zhu2019adapting} and registration \cite{yuan2021self,lu2021hregnet,zeng2021corrnet3d,chen2020unsupervised}. Due to its unordered nature, the network for processing the point cloud is less mature than 2D images. Various network architectures have been proposed for learning with point clouds. They mainly focus on designing mechanisms for aggregating neighbourhood information. PointNet based methods \cite{qi2017pointnet,qi2017pointnet++,wu2019pointconv} mainly apply multi-layer perception networks directly on the coordinates of input point cloud for feature extraction and cropping operation is usually employed for local feature aggregation. Sparse convolution-based methods \cite{graham2015sparse,su2018splatnet,choy20194d} mainly represent the 3D point cloud with sparse rectangles or permutohedral lattice and define 3D convolution kernels accordingly. In our work, we use the MinkowskiNet \cite{choy20194d}, a kind of sparse convolutional network, with U-Net like architecture as the backbone for learning point cloud representations. 

\paragraph{\textbf{Transfer Learning in 3D}}
Transfer learning has been widely employed in various deep learning-based tasks. The main purpose is to improve neural networks' performance and generalization ability under limited annotated data. In 3D scenarios, transfer learning becomes much more critical due to the difficulty of acquiring 3D labelled data. Generally, deep transfer learning include but not limited to three categories: Self-supervised learning that pre-train the network with extra dataset, and fine-tune on the downstream tasks \cite{yosinski2014transferable,mahajan2018exploring,xie2020pointcontrast,hou2021exploring}; Domain adaptation between the source domain and target domain \cite{saenko2010adapting,wang2020alleviating,qin2019pointdan,tzeng2017adversarial,cui2021structure}; and few-shot/semi-supervised learning with few annotated data \cite{yu2020transmatch,zhao2021few,wang20213dioumatch,jiang2021guided,chen2022semi}. Compared with the above transfer learning methods, our problem setting refers to 3D synthetic models for supervision and inferring the objects on 3D real scenes. Thus we name it Referring Transfer Learning. Besides, there are some methods \cite{yi2019gspn,avetisyan2019end,gupta2015aligning} leverages synthetic objects for learning in real scenes. However, they require scene annotation for supervision. While our method only learns from the labelled synthetic models. Some optimization-based methods \cite{knopp2011scene,lai2010object,kimacquisition,ishimtsev2020cad,song2014sliding,litany2017asist,li2015database,nan2012search} that fit 3D synthetic models to the real scene's objects for reconstruction, object replacement and registration are out of the scope of our intention. Unlike the above methods, we study the neural network's model-to-scene and synthetic-to-real generalization ability, which transfer knowledge from 3D synthetic models to real scenes for the 3D scene understanding.

\paragraph{\textbf{3D Data Augmentation}}
Data augmentation, as a fundamental way for enlarging the quantity and diversity of training datasets, plays an important role in various deep learning tasks. It is especially important in the 3D deep learning scenario, which is notoriously data hungry. Simple data augmentation schemes like random rotation, translation, jittering, scaling, and cropping are commonly used in point cloud deep learning methods. Such simple techniques can usually achieve much more performance gain compared with complex network architecture designs. Recently, several attempts have been made on designing new 3D data augmentation schemes and studying 3D data augmentation techniques in systematic ways \cite{li2020pointaugment,chen2020pointmixup,zhang2017mixup,zhang2021pointcutmix,nekrasov2021mix3d}. PointAugment \cite{li2020pointaugment} proposes a learnable point cloud augmentation module to make the augmented data distribution better fit with the classifier. PointMixup \cite{chen2020pointmixup} extends Mixup \cite{zhang2017mixup} augments the data by interpolating between data samples. PointCutMix \cite{zhang2021pointcutmix} further extend Mixup strategy and perform mixup on part level. Mix3D \cite{nekrasov2021mix3d} creates new training samples by combining two augmented scenes. Inspired by the above methods, we utilize data augmentation strategies for synthetic models to cover the diversity of the objects in real unseen scenes.

\paragraph{\textbf{Memory Networks}}
The Prototype-based Memory network has been applied to various problems. NTM \cite{graves2014neural} introduces an attention-based memory module to improve the generalization ability of the network. Gong et al. \cite{gong2019memorizing} adopt a memory augmented network to detect the anomaly. Prototypical Network \cite{snell2017prototypical} utilize category-specific memories for few-shot classification. Liu et al. \cite{liu2019large} and He et al. \cite{he2020learning} solve the long-tail issue with the prototypes. In this paper, we adopt the learnable prototypes as the support points to formulate a convex hull that alleviates the domain gap between the synthetic model and the real scene.

\section{Referring Transfer Learning}
\paragraph{\textbf{Problem Definition}}
Give 3D synthetic models $\{M_{i}\}_{i=1}^{N^m}$ with labels $\{G_{i}\}_{i=1}^{N^m}$ as reference, the neural network aims to recognize specific categories of objects on real scene scans $\{S_{j}\}_{j=1}^{N^s}$. $N^m$ and $N^s$ are the number of model and scene, respectively. Note that we use unlabelled scenes and labelled synthetic models for training the network. We name it Referring Transfer Learning (RTL) as it refers to 3D synthetic models for supervision and transfers knowledge from synthetic models to real scenes. The main challenge is solving the model-to-scene (from a single model to the scene) and synthetic-to-real (from synthetic model to real scene’s object) gap between the synthetic model and the real scene. RTL in this paper is instantiated as a point-wise classification problem (semantic segmentation), \ie, predicting the possibility of each point that belongs to specific classes. 

\begin{figure*}
  \centerline{\includegraphics[width=1\textwidth]{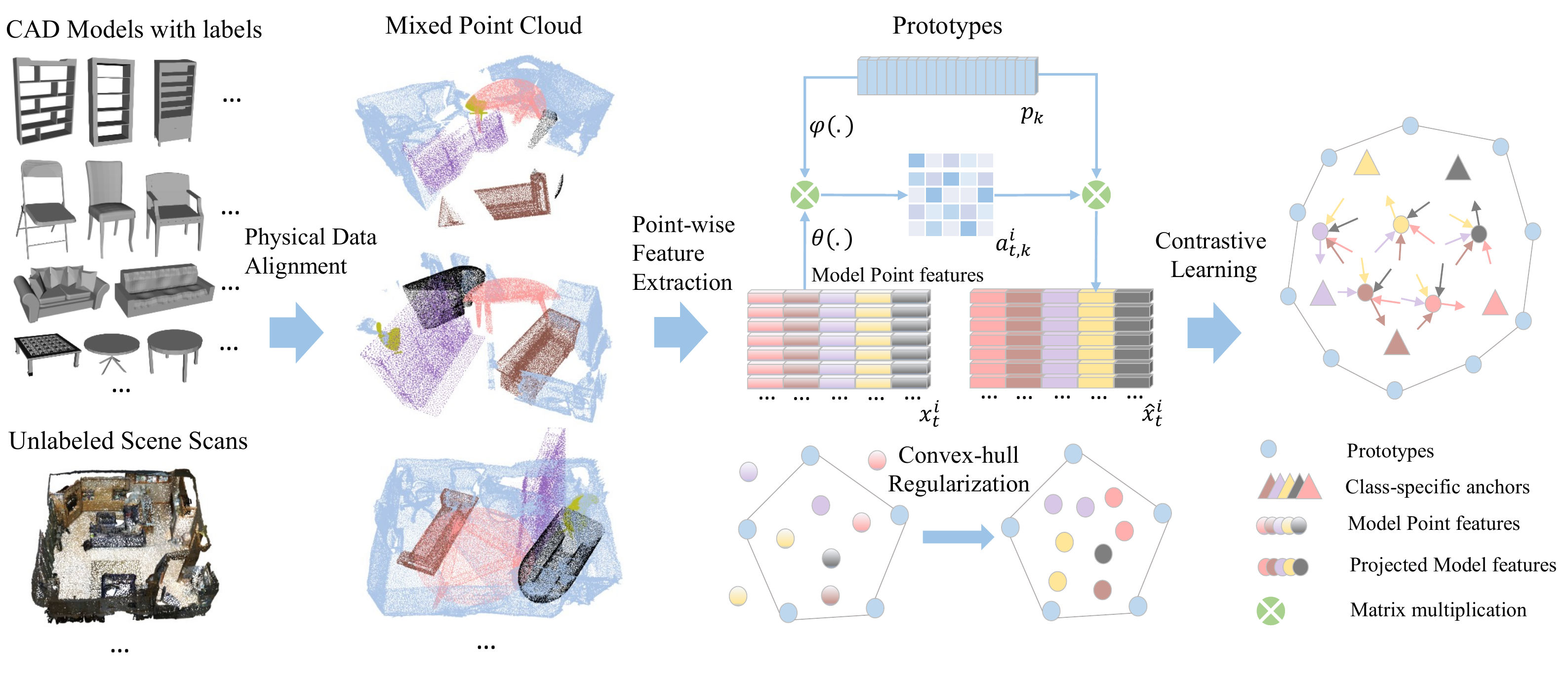}}
  \caption{The framework of our method in the training stage. Firstly, the synthetic models are cover the diversity of the real scene's objects via physical data alignment, including random rotation, scaling, cropping, and mixing up with the scene (after mixing up, these coloured points are from synthetic models). Secondly, we extract the point-wise feature from the mixed point cloud and map the point features into a convex hull space surrounded by a group of learned prototypes. In the end, the aligned point features of models are clustering with the class-specific anchors in metric space.}
  \label{fig:framework}
  \vspace{-2ex}
\end{figure*}

\paragraph{\textbf{Approach Overview}}
As illustrated in Fig. \ref{fig:framework}, our method consists of two alignment operations. Firstly, we physically align the synthetic models to cover the diversity of the objects in a real scene scan using data processing techniques. Secondly, a novel convex-hull regularized feature alignment is proposed, in which we map the point features into a unified convex hull space surrounded by a group of learned prototypes. In the end, we perform contrastive learning on the mapping features. The cooperation of these steps leads to the success of inference on a real unseen scene by learning from the synthetic models. In what follows, we will present these components in detail.

\subsection{Physical Data Alignment}
Physical-level alignment is an intuitive and straightforward option to ease the model-to-scene difficulty. Inspired by current data augmentation methods, a series of data processing approaches are introduced to cope with the synthetic models to cover the diversity of the objects in a real scene scan, including \textbf{1)} point sampling, scaling, rotation, cropping and \textbf{2)} point mixing.

\paragraph{\textbf{Point Sampling, Scaling, Rotation and Cropping}}
CAD models are presented in Mesh format that consists of the vertexes and faces. To unify the data format, we transfer the model mesh to a uniform point cloud by Poisson Disk Sampling \cite{yuksel2015sample}, and ensure the density closer to the scene scan. For consistent local feature extraction, We roughly scale the size of the synthetic models to match the real scene's object size with prior knowledge (not model fitting). Besides, random rotation transformation is also applied to capture the pose diversity of an object. Finally, considering that the object in a scene scan is always partial observed, a random cropping strategy is designed to simulate this scenario. Specifically, we first randomly sample 2$\thicksim$5 points from the model as anchor points and then cluster all points based on their Euclidean distance to the anchor points. During training, one of a cluster will be randomly filtered.

\paragraph{\textbf{Point Mixing}}
Unlike the synthetic model, the point feature of the object in a real scene scan is always affected by the surrounding object. Therefore, we mix up with the synthetic models and unlabeled scenes to alleviate the difference. Specifically, we randomly place the synthetic models on the scene floor (regardless of the layout), with or without filtering the overlapped points randomly.

\begin{figure}
  \centerline{\includegraphics[width=0.9\textwidth]{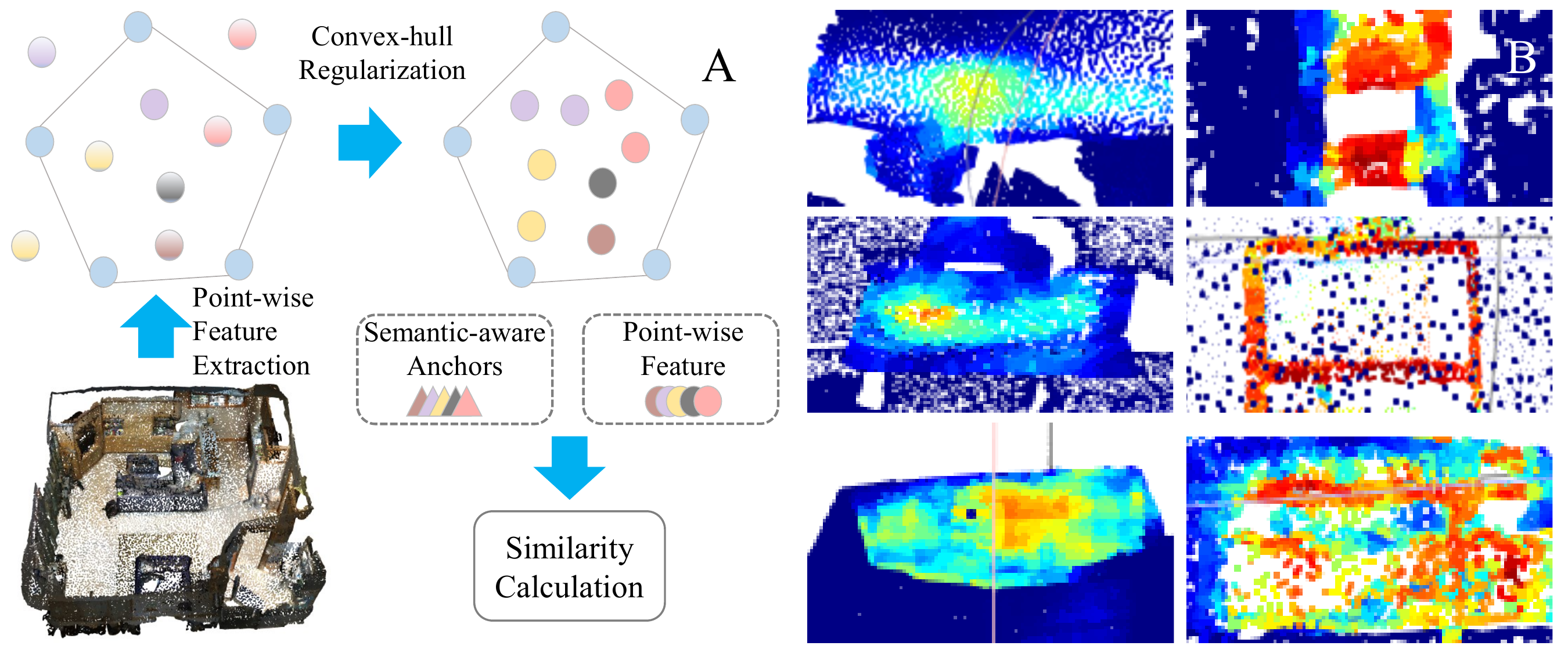}}
  \caption{The sub-picture A is the framework in the inference stage. The sub-picture B is the visualization of the activation region for two learned prototypes (left column for plane structure and right column for the hole).}
  \label{fig:visual_sturcture}
  \vspace{-3ex}
\end{figure}

\subsection{Convex-hull Regularized Feature Alignment}
Since the network is only trained on the synthetic models, its feature space typically differs from the object in a real scene scan, leading to low inferring accuracy. Inspired by the convex hull theory~\cite{rockafellar2015convex}, we propose a novel module to project point features into a convex hull space via the learnable prototypes for promoting the alignment. In the following, we revisit the convex hull and describe how prototypes are used for feature alignment.

\paragraph{\textbf{Revisiting Convex Hull Theory}}
Convex hull is a fundamental concept in computational geometry. It is defined as the set of all convex combinations of the support points, where the convex combination is a linear combination and all coefficients are non-negative and sum to 1. By definition, if a point is presented as a convex combination of the support points, it must remain in the convex hull. In this way, the distributions from two different domains can be constrained to unified feature space. Therefore, we project all features into a convex hull to alleviate the domain gap.

\paragraph{\textbf{Formulation}}
We set a group of learnable prototypes $\{p_k\}_{k=1}^{K}$ with $p_k \in \mathbb{R}^{D}$ and $K > D$, where $K$ denotes the number of prototypes, and $D$ is the dimension of a prototype. Note that prototypes are directly learned from the synthetic models and updated via back-propagation. The prototypes are designed to learn the base structural elements of 3D models (Fig. \ref{fig:visual_sturcture} (B)). In this context, a point feature can be reconstructed by a linear combination of the related structural elements. Given the point features $\{x_t^i\}_{t=1}^{T}$ with $x_t^i \in \mathbb{R}^{D}$ extracted by the encoder $E$ from the $i$-th synthetic model with $T$ points, the corresponding mapping feature $\{\hat{x}_t^i\}_{t=1}^{T}$ is obtained by the following function.

\begin{equation}\label{equ:featureMapping}
\hat{x}_{t}^i = \sum^{K}_{k=1} a^{i}_{t,k}*p_{k}, \sum^{K}_{k=1} a^{i}_{t,k} = 1,
\end{equation}
where $a^{i}_{t,k}$ serves as the coefficient to the corresponding prototype, defined by:

\begin{equation}\label{equ:coefficient}
\begin{split}
a^{i}_{t,k} &= \exp(\lambda*d(\theta(x_t^i), \varphi(p_{k})))/\Gamma,\\ \Gamma&=\sum^{K}_{k=1}\exp(\lambda*d(\theta(x_t^i), \varphi(p_{k}))),
\end{split}
\end{equation}
where $d(\cdot)$ measures the similarity between the point feature and the $k$-th prototype, which is dot product operation in this work. $\theta(\cdot)$ and $\varphi(\cdot)$ denote the key and the query function \cite{vaswani2017attention}, respectively. $\lambda$ is the inversed temperature term \cite{chorowski2015attention}.

Essentially, the feature embedding $\hat{x}_{t}$ is a convex combination of the prototypes, where the coefficients $a^{i}_{t,k} > 0$ and sum to 1. Therefore, $x_{t}$ is mapped into a convex $\mathbb{W} \subset \mathbb{R}^{D}$, where $\mathbb{W}$ is a closure and compact metric space surrounded by the learned prototypes.

In the inferring phase (Fig. \ref{fig:visual_sturcture} (A)), the point feature $x_{t} \in \mathbb{R}^{D}$ from a real unseen scan first accesses the most relevant prototypes to obtain the coefficients. Then, it is transferred to a mapping feature $\hat{x}_{t} \in \mathbb{W}$ which is the convex combination of the prototypes. In this way, the feature space of the synthetic models and the objects in an real scene are projected to the unified subspace $\mathbb{W}$, leading to the network a better generalization ability.

\subsection{Contrastive Learning in Metric Space}

As all point features are projected to the subspace $\mathbb{W}$, we cluster these points to ensure they are inner-class compact and inter-class distinguishable. We set the class-specific anchors $\{h_c\}_{c=1}^{C}$ with $h_c \in \mathbb{R}^{D}$ to indicate the clustering centers in metric space, which include $C - 1$ foreground classes and one background class. Given the point features $\{\hat{x}_t^i\}_{i=1,t=1}^{N^m,T}$ from $N^m$ synthetic models$\{M_i\}_{i=1}^{N^m}$, we pull in whose points to the corresponding anchor while pushing away from the rest of anchors according to their semantic labels $\{G_i\}_{i=1}^{N^m}$. Therefore, the points are compact for inner class and distinguishable for inter classes in the metric space $\mathbb{W}$. For simplicity, we adopt Cross-Entropy loss in this paper. Therefore, the objective function $\mathcal{L}$ is as follows.
\begin{equation}\label{equ:lossFunction}
\mathcal{L} = -\sum_{i=1}^{N^m}\sum_{t=1}^{T} d(\hat{x}_{t}^{i}, h_{G_i}) + \sum_{i=1}^{N^m}\sum_{t=1}^{T} \log(\sum_{c=1}^{C} \exp(d(\hat{x}_{t}^{i}, h_c))).
\end{equation}

When inferring an unseen scene scan $S_j$ with the point features $\{\hat{x}_t^j\}_{i=1}^{T}$, the possibility distribution of the $t$-th point that belongs to each class is determined by the similarities between the point feature $\hat{x}_t^j$ and the class-specific anchors $\{h_c\}_{c=1}^{C}$ (Fig \ref{fig:visual_sturcture} (A)).

\begin{equation}\label{equ:inferring}
l_t^c = \exp(d(\hat{x}_t^j, h_{c}))/\gamma, \gamma=\sum^{C}_{c=1}\exp(d(\hat{x}_t^j, h_{c})),
\end{equation}
where $l_t^c$ is the possibility that the $t$-th point belongs to the $c$-th class, $\gamma$ is a normalization term.

\section{Experiments}
\subsection{Dataset}
We conduct the experiments on ModelNet40 \cite{wu20153d}, ScanNet \cite{dai2017scannet} and S3DIS \cite{2017arXiv170201105A}  datasets, where ModelNet provides the synthetic model for referring, and the scene scans in ScanNet and S3DIS are for evaluation.

\vspace{-1ex} 
\paragraph{\textbf{ModelNet}}
is a comprehensive clean collection of 3D CAD models for objects, composed of 9843 training models and 2468 testing models in 40 classes. We transfer the model mesh to 8196 uniform points by Poisson disk sampling \cite{yuksel2015sample}. We take the 9843 models from the training set as the synthetic models in our method.

\vspace{-1ex} 
\paragraph{\textbf{ScanNet}}
contains 1603 scans, where 1201 scans for training, 312 scans for validation and 100 scans for testing. The 100 testing scans are used for the benchmark, and their labels are unaccessible. There are 11 identical classes to the ModelNet40 dataset, including the chair, table, desk, bed, bookshelf, sofa, sink, bathtub, toilet, door, and curtain. We take the 1201 scans to mix up with the synthetic models for training (labels are not used), and the rest of the 312 scans are used to evaluate the performance. 

\vspace{-1ex} 
\paragraph{\textbf{S3DIS}}
consists of 271 point cloud scenes across six areas for indoor scene semantic segmentation. There are 13 categories in the point-wise annotations, where four identical classes to the ModelNet40 dataset, including the chair, table, bookcase (bookshelf) and sofa. We utilize Area 5 as the validation set and use the other five areas as the training set (labels are not used), the same with the previous works~\cite{li2018pointcnn,qi2017pointnet,jiang2021guided}

\subsection{Evaluation Metric}
The goal is to detect the objects (point clouds) that belong to the identical class with synthetic models. Therefore, we calculate the class-specific point-wise possibility on the scene scan and adopt the mean Average Precision (mAP) to measure the performance for each class. AmAP is the average mAP of all classes. We believe AmAP is more suitable than mIoU because the foreground category of objects occupies a small proportion in a whole scene.

\begin{table*}[h]
	\centering
	\caption{Evaluation on the ScanNet. MinkUNnet is the baseline method. We apply point-wise feature adaptation in ADDA and instance adaptation in ADDA$\dagger$. 'Supervised' indicates training with annotated scenes. }\label{tab:Scannet}
	\scalebox{0.82}{
	\begin{tabular}{c c c c c c c c c c c c c}
        \hline
        \multirow{2}*{Method} &
		\multirow{2}*{AmAP} &
		\multirow{2}*{chair} &
		\multirow{2}*{table} &
		\multirow{2}*{bed} &
		\multirow{2}*{sink} &
		\multirow{2}*{bathtub} &
		\multirow{2}*{door} &
		\multirow{2}*{curtain} &
		\multirow{2}*{desk} &
		\multirow{2}*{bookshelf} &
		\multirow{2}*{sofa} &
		\multirow{2}*{toilet} \\

		~ & ~ \\
	    \hline
	    Baseline & 10.93 & 8.09 & 10.60 & 15.97 & 2.53 & 12.40 & 9.77 & 13.92 & 4.66 & 26.76 & 11.49 & 4.07\\
        ADDA \cite{tzeng2017adversarial} & 28.93 & 29.93 & 40.88 & 36.82 & 8.03 & 31.60 & 13.59 & 25.10 & 20.01 & 35.49 & 33.90 & 42.92 \\
        ADDA$\dagger$ \cite{tzeng2017adversarial} & 38.14 & 67.03 & \textbf{48.60} & 29.77 & 20.36 & 29.70 & 12.04 & 23.06 & 26.57 & 42.69 & 51.41 & 68.31 \\
	    PointDAN \cite{qin2019pointdan} & 32.92 & 58.31 & 40.39 & 20.96 & 12.58 & 31.65  & 11.79 & 17.65 & 26.04 & 48.31 & 41.25 & 53.18 \\
	    3DIoUMatch \cite{wang20213dioumatch} & 42.25 & 63.34 & 41.24 & 43.58 & 26.53 & \textbf{46.61} & 15.52 & 24.45 & 26.75  & 45.49 & 54.49 & \textbf{76.81} \\
	    Ours & \textbf{46.08} & \textbf{67.19}& 46.73 & \textbf{45.65}& \textbf{31.28} & 43.36 & \textbf{17.47} & \textbf{34.43} & \textbf{29.15} & \textbf{59.37} & \textbf{60.48} & 71.77\\
        \hline
	    Supervised & 78.98  & 91.44 & 72.89 & 76.35 & 83.94 & 84.84  & 58.47 & 74.22 & 72.96 & 75.76 & 84.81 & 93.10 \\
	    Supervised+ours & 79.46  & 92.16  & 74.37  & 76.50  & 85.18 & 85.09 & 56.57 & 70.68 & 73.54 & 77.03 & 87.44 & 95.52\\

	\end{tabular}}
\end{table*}

\begin{table}[h]
    \vspace{1ex}
	\centering
	\caption{Evaluation on the S3DIS area 5 test. MinkUNnet is the baseline method. We apply point-wise feature adaptation in ADDA and instance adaptation in ADDA$\dagger$. 'Supervised' indicates training with annotated scenes. }\label{tab:S3DIS}
	\scalebox{1}{
	\begin{tabular}{c c c c c c}
        \hline
        \multirow{2}*{Method} &
		\multirow{2}*{AmAP} &
		\multirow{2}*{chair} &
		\multirow{2}*{bookshelf} &
		\multirow{2}*{sofa} &
		\multirow{2}*{table} \\

		~ & ~ \\
	    \hline
	    Baseline & 15.65 & 3.75 & 37.72 & 13.34 & 7.80\\
        ADDA \cite{tzeng2017adversarial} & 26.57  & 26.67& 54.70 & 15.91 & 9.00 \\
        ADDA$\dagger$ \cite{tzeng2017adversarial} & 30.04  & 57.80 & 54.70 & 22.91 & 8.70 \\
	    PointDAN \cite{qin2019pointdan} & 39.80 & 56.60& 52.34 & 32.06 & 18.20 \\
	    3DIoUMatch \cite{wang20213dioumatch} & 49.60 & 69.07 & 56.87 & \textbf{54.25} & 18.21 \\
	    Ours & \textbf{55.49} & \textbf{70.86} & \textbf{62.68} & 47.87 & \textbf{40.57} \\
        \hline
	    Supervised & 90.44 & 97.05 & 87.84 & 86.83  & 90.06  \\
	    Supervised+ours & 92.57 & 97.83 & 89.86 & 92.05 &  90.52 \\

	\end{tabular}}
\end{table}

\subsection{Implementation Details}
We adopt MinkowskiNet14 \cite{choy20194d} as the backbone to extract the point-wise feature. Thus, the feature dimension $D$ is set to be 96. The key $\theta(\cdot)$ and the query $\varphi(\cdot)$ function are linear transformation and output 16-dimensional vectors. The voxel size of all experiments is set to be 5 cm for efficient training. Our method is built on the Pytorch platform, optimized by Adam with the default configuration. The batch size for the ModelNet,  ScanNet and S3DIS are 4 * ($Q$ + 1), 4 and 4, respectively, indicating that one scan is mixed up with ($Q$ +1) synthetic models, where $Q$ is the number of identical classes between two datasets and one for the negative sample. Since there is no colour in the synthetic models, we set the feature in the ScanNet and S3DIS dataset to be a fixed tensor (1), identical to that in the ModelNet. Training 200 epochs costs 15 hours on two RTX 2080 TI GPUs. During training, we randomly rotate the models and scans along the z-axis, randomly scale the model and scene with scaling factor 0.9-1.1 and randomly displace the model's location within the scene. If there are overlapped points, we randomly filter or maintain them. We take all identical classes as foreground classes to evaluate the performance and utilize the remaining classes as negative samples for contrastive learning. The detailed network configurations are included in supplementary materials.

\subsection{Results and Discussions}
In this section, we report the performance of two tasks: 1. learning from the synthetic models and inferring in a real scene on ScanNet and S3DIS datasets; 2. fine-tuning the pre-trained network with labelled scene scans for semantic segmentation and object detection. In the end, we discuss the limitations and potential directions for future works.

\begin{table}[h]
	\centering
	\caption{Fine-tuning on the Scannet and S3DIS dataset for semantic segmentation task. We omit the \% to show the mIoU performance. The number in () donates the improved accuracy compared with purely supervised training.}\label{tab:fineTuning_scannet}
	\vspace{-1ex}
	\scalebox{0.95}{
	\begin{tabular}{l|c c c | c c}
		\hline
		\multirow{2}*{\tabincell{c}{Model}} & \multicolumn{3}{c}{Scannet} &\multicolumn{2}{c}{S3DIS} \\
		\cline{2-6}
		~&5\%  data & 10\% data&100\%  data & MinkNet14 & MinkNet34\\

        \hline
		Trained from scratch  & 50.24 & 54.86 & 63.05 & 56.44 & 58.63 \\
		PointContrast \cite{xie2020pointcontrast}  & 55.31(5.07) & 58.68(3.82) & \textbf{65.03(1.98)} & 58.65(2.21) & \textbf{60.71(2.08)} \\
		Ours & \textbf{56.12(5.88)} & \textbf{59.03(4.17)} & 64.93(1.88) & \textbf{58.89(2.63)} & 60.68(2.05)\\

	\end{tabular}}
	\vspace{-1ex}
\end{table}

\begin{table}[h]
	\centering
	\vspace{-1ex}
	\caption{Fine-tuning on the Scannet for 3D object detection results. We omit the \% to show the mAP performance. The number in () donates the improved accuracy compared with purely supervised training.}\label{tab:fineTuning_scannet_detection}
	\scalebox{1}{
	\begin{tabular}{l|c c}
		\hline
		Model& mAP@0.5 & mAP@0.25 \\
		\hline
		Trained from scratch & 31.82 & 53.39\\
		PointContrast~\cite{xie2020pointcontrast} & 34.30(2.48) & \textbf{55.56(2.17)}\\
		Ours & \textbf{34.46(2.64)} & 55.54(2.15) \\
	\end{tabular}}
	\vspace{-3ex}
\end{table}

\subsubsection{Learning From the Synthetic Models and Inferring in an Real Scene}
\begin{figure*}
  \centerline{\includegraphics[width=1\textwidth]{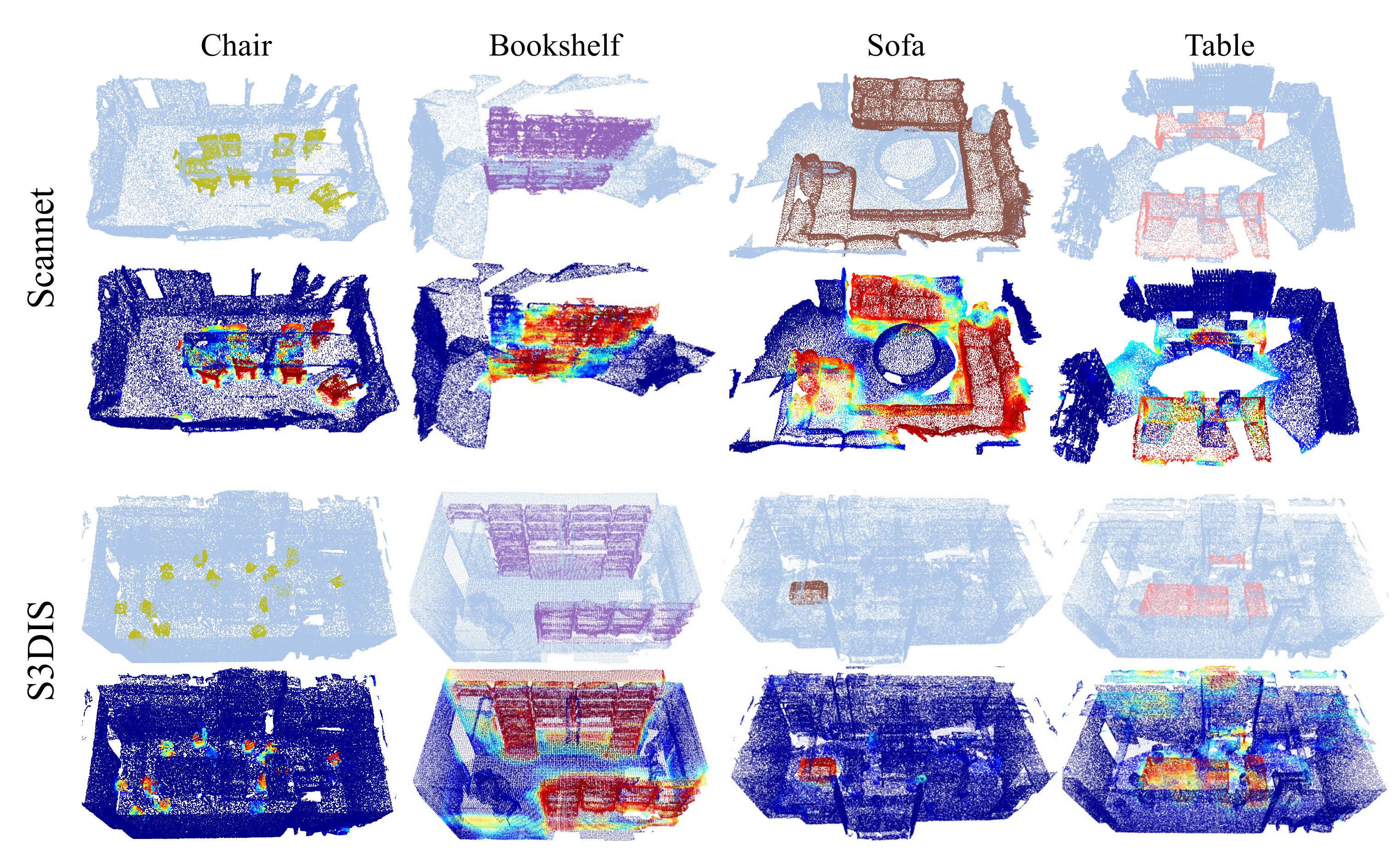}}
  \vspace*{-1.5ex}
  \caption{Visualization of inferring on real scene scans by our method. The first two rows are from the ScanNet dataset, while the last two rows are from the S3DIS dataset. We show the pairs of ground truth (up) and the inferring results (down). From left to right are chair, bookshelf, sofa and table, respectively.}
  \label{fig:visual}
  \vspace{-2ex}
\end{figure*}

\paragraph{\textbf{Baselines}}
\vspace{-3ex}
There are no deep learning-based methods investigated on this problem to the best of our knowledge. To build the baseline method (baseline in Table \ref{tab:S3DIS} and Table \ref{tab:Scannet}) for comparison, we first resize the synthetic models to the same scale with the scene's objects, then extract the point feature for individual models and classify the points with the model labels. During the inferring phase, we directly apply the trained network on the scene scan for point-wise classification. Note that it does not consider the model-to-scene and synthetic-to-real gaps between the 3D synthetic models and the real scene scan. We also compared our method with a semi-supervised method (3DIoUMatch \cite{wang20213dioumatch}) and two unsupervised domain adaptation methods (ADDA \cite{tzeng2017adversarial} and PointDAN \cite{qin2019pointdan}). Note that they are failed if directly learning from the synthetic models. For fair compared with convex-hull regularized feature alignment, the mixed point cloud (Fig. \ref{fig:framework}) is regarded as labelled data in 3DIoUMatch, and as the source domain in ADDA and PointDAN. To adapt 3DIoUMatch for the semantic segmentation task, we calculate the mask IoU instead of the bounding box IoU.
\vspace{-2ex} 
\paragraph{\textbf{ScanNet}}
As shown in Table \ref{tab:Scannet}, our method achieves 46.08\% AmAP to identify 11 types of objects on the ScanNet validation dataset, which significantly outperforms other methods. Compared with the baseline, the improvements for each class are from 8\%-67\%, indicating the effectiveness of the whole framework. Furthermore, the convex-hull regularized feature alignment is verified to be feasible as it achieves a better performance than 3DIoUMatch, ADDA and PointDAN. We also show the performance by training on the annotated scene scans (Supervised), which is 78.98\% AmAP, indicating that there is still much room for improvement. When training on both synthetic models and annotated scene scans (supervised+Ours), the performance is higher than that only training on ground truth. The qualitative evaluation is shown in Fig. \ref{fig:visual}, indicating the network is well adaptive to the scenes by learning the synthetic models. More results are in supplementary materials.

\vspace{-2ex} 
\paragraph{\textbf{S3DIS}}
Our method achieves 55.49\% AmAP for identifying four classes on the S3DIS dataset (Table \ref{tab:S3DIS}). The AmAP improvement is 40\% compared with the baseline method and is 6\%-29\% compared with other methods, showing the effectiveness of the framework and the convex-hull regularized feature alignment. Besides, the performance of training on both the labelled scene scans and the synthetic models (Supervised+ours) is better than the counterpart (Supervised), indicating that our method favour the performance of purely supervised learning. The qualitative evaluation is shown in Fig. \ref{fig:visual} and supplementary materials.

\vspace{-2ex} 
\subsubsection{Fine-tuning in the Downstream Tasks}
Our method is beneficial for the downstream tasks. We first pre-train the network by the synthetic models and then fine-tune the network with different proportions of labelled scans for scene segmentation on scannet, where mIoU is adopted as the evaluation metric. For the S3DIS dataset, we evaluated with two backbone networks (MinkowskiNet14 and MinkowskiNet34). As shown in Table \ref{tab:fineTuning_scannet}, our method is comparable with state-of-the-art method Pointcontrast \cite{xie2020pointcontrast}. Note that we compare the original version of Pointcontrast without leveraging additional models. Compared with purely supervised counterparts, a significant improvement could be observed in the two datasets for both seen and unseen categories (not shown on synthetic models). Besides, our method is also beneficial for object detection task (Table \ref{tab:fineTuning_scannet_detection}). It is because the network learns meaningful features from the synthetic models. Another potential reason is that the synthetic models alleviate the long-tail distribution issue. More experiment results are in supplementary materials.

\begin{table*}[h]
	\centering
	\caption{Ablation experiments. MinkUNnet is the baseline method. PDA and FA donate physical data alignment and convex-hull regularized feature alignment, respectively.}\label{tab:ablation}
	\scalebox{0.74}{
	\begin{tabular}{c c c c c c c c c c c c c}
        \hline
        \multirow{2}*{Method} &
		\multirow{2}*{AmAP} &
		\multirow{2}*{chair} &
		\multirow{2}*{table} &
		\multirow{2}*{bed} &
		\multirow{2}*{sink} &
		\multirow{2}*{bathtub} &
		\multirow{2}*{door} &
		\multirow{2}*{curtain} &
		\multirow{2}*{desk} &
		\multirow{2}*{bookshelf} &
		\multirow{2}*{sofa} &
		\multirow{2}*{toilet} \\

		~ & ~ \\
	    \hline
	    MinkUNet & 10.93 & 8.09 & 10.60 & 15.97 & 2.53 & 12.40 & 9.77 & 13.92 & 4.66 & 26.76 & 11.49 & 4.07\\
	    \hline
	    MinkUNet+PDA$_{noDA}$ & 23.44 & 36.90 & 35.57 & 26.30 & 11.98 & 17.29 & 11.01 & 21.07 & 11.58 & 25.03 & 16.40 & 44.66 \\
	    MinkUNet+PDA$_{negaSc}$ & 23.04  & 41.74 & 25.05 & 21.92 & 7.86  & 23.01 & 13.88 & 21.93 & 14.79 & 23.80 & 13.08 & 46.39 \\
	    MinkUNet+PDA$_{negaMo}$ & 41.34 & \textbf{73.12} & 40.05 & 42.73 & 29.17  & 19.97 & 17.52 & 23.33 & 33.40 & 50.34 & 59.36 & 65.71 \\
	    MinkUNet+PDA$_{negaMo\_noCo}$ & 37.83 & 72.62 & 39.61 & 37.49  & 14.95 & 16.93 & 18.33 & 21.29 & 32.84 & 46.09 & 59.44 & 56.53 \\
	    
	    MinkUNet+PDA & 42.19 & 67.53 & 47.64 & 41.82  & 27.01  & 32.47  & 13.68  & 27.82  & 26.78  & 50.54  & 58.75  & 70.07 \\
        \hline
        MinkUNet+PDA+FA$_{K64}$ & 43.16 & 67.37  & \textbf{48.27}  & 41.51  & 20.38 & 33.90 & 16.30  & 31.55  & 27.17  & 52.32  & \textbf{61.19}  & 74.84  \\
	    MinkUNet+PDA+FA$_{K128}$ & \textbf{46.08} & 67.19& 46.73 & 45.65& \textbf{31.28} & 43.36 & 17.47 & 34.43 & 29.15 & \textbf{59.37} & 60.48 & 71.77\\
        MinkUNet+PDA+FA$_{K256}$ & 43.81 & 68.15  & 45.12 & \textbf{51.08} & 20.12 & 38.14 & 17.10 & 25.64 & \textbf{33.99} & 48.23 & 59.32  & 75.05 \\
	    MinkUNet+PDA+FA$_{T4}$ & 44.51 & 69.05 & 47.28 & 49.72 & 18.26  & \textbf{45.30} & \textbf{18.46} & 28.83 & 26.16 & 49.94 & 57.23 & \textbf{79.40} \\
	    MinkUNet+PDA+FA$_{T0.1}$ & 43.79 & 70.47 & 42.44 & 44.26 & 20.95 & 39.57 & 16.00 & 29.92 & 27.65 & 52.04 & 59.14  & 79.20\\
	    MinkUNet+PDA+FA$_{cos}$  & 43.45  & 69.16  & 45.84 & 42.53  & 21.47 & 37.38 & 15.36 & 29.20 & 29.06 & 51.72 & 59.16 & 77.06\\

	\end{tabular}}
	\vspace{-2ex}
\end{table*}

\vspace{-2ex} 
\subsubsection{Limitations and Future Work} 
There are some limitations in the current implementation of Referring Transfer Learning. Firstly, the synthetic objects are randomly placed, and layout information is not considered. Secondly, only synthetic 3D models are used for training, which only offers 3D geometry features. Thus, other modality information may provide supplementary information for more accurate perception in the real world, such as the colour information by images. Besides, by leveraging word embedding as auxiliary information, the network could recognize the unseen categories that are not available in the synthetic models. Inspired by the current zero-shot segmentation method \cite{michele2021generative,zhang2021prototypical,bucher2019zero}, we implemented a baseline method in the supplementary materials. We will share the code and data with the community for future works.

\subsection{Ablation study}
We conduct experiments on ScanNet to verify the effectiveness of different components in our method, including physical data alignment (PDA), and convex-hull regularized feature alignment (FA). In the following, we present the configuration details and give more insights into what factors affect the performance.

\paragraph{\textbf{Effect of Physical Data Alignment}}
MinkUNet+PDA donates the baseline with physical data alignment. Observing from (MinkUNet, MinkUNet+PDA), AmAP is improved by 31.26\%. By physical aligning the models and the scene scans, the model is closer to the real scene's objects and the performance is improved accordingly. We dig into PDA by exploring the following configurations. 

Firstly, we investigate how data augmentation (DA) influences performance, including random scaling and rotation. These operations cover the diversities of the scene's objects. As shown in Table \ref{tab:ablation}, the performance greatly reduced if without applying DA (MinkUNet+PDA$_{noDA}$ (23.44 AmAP) VS MinkUNet+PDA (42.19 AmAP)). Besides, we find the random cropping is beneficial for performance improvement due to the real scene's objects are often partially scanned (MinkUNet+PDA$_{negaMo\_noCo}$ (37.83 AmAP) is without random cropping). To summary, suitable data alignment that makes synthetic models more realistic is critical for the network to learn general features with domain-invariant property.

Secondly, to explore how to conduct negative samples, we try to take the scene's points as the negative samples for contrastive learning (Mink+PDA$_{negaSc}$). We find that the performance (23.04 AmAP) is significantly worse than Mink+PDA (42.19 AmAP), which uses the synthetic models as negative samples. It is probably that the network learns the artefacts to distinguish the synthetic models from the scene. The artefacts are mainly caused by the mixing up operation, such as the overlapped/disjointed points. As a result, the network could not generalize the knowledge to a clear scene without such artefacts. However, by contrastive learning on the positive and negative synthetic models with artefacts, the network forces to learn geometric features for inferring the real scene.

Lastly, to understand the role of mixing the scene, we only mix up the synthetic models together and exclude the scene points (MinkUNet+PDA$_{negaMo}$). Surprisingly, the performance is comparable with the counterpart MinkUNet+PDA (41.34 AmAP VS 42.19 AmAP). It shows that the realistic background may not critical for the network to infer objects in an unseen scene.

\paragraph{\textbf{Effect of Feature Alignment}}
Since feature domain gaps exist between the synthetic models and the objects in a real scan, we use the prototypes to align their features into an unified feature space (MinkUNet+PDA+FA). The experiment shows that the improvement is about 4\% for AmAP (MinkUNet+PDA VS MinkUNet+PDA+FA${_K128}$). The inversed temperature $\lambda$ and the number of prototypes $K$ are two hyper-parameters for the feature alignment module. $\lambda$ indicates the smoothness of the coefficient distribution and can be regarded as a regularization term to prevent network degradations. We present the results when $\lambda$ is 0.1, 0.5 and 4, respectively. (MinkUNet+PDA+FA$_{T0.1}$, MinkUNet+PDA+FA$_{K128}$ and MinkUNet+PDA+FA$_{T4}$). We choose $\lambda$ to be 0.5 empirically. The number of prototypes $K$ is another hyper-parameter. We respectively evaluate 
it with the configurations MinkUNet+PDA+FA$_{K64}$, MinkUNet+PDA+FA$_{K128}$, and MinkUNet+PDA+FA$_{K256}$. The network achieves the best performance when $K$ is set to be 128. Besides, we show the result when the key $\theta(\cdot)$ and the query function $\varphi(\cdot)$ are identity mapping function in the setting MinkUNet+PDA+FA$_{cos}$. The performance is slightly worse than the full method MinkUNet+PDA+FA$_{K128}$.

\section{Conclusion}
We formulate and investigate the problem of transferring knowledge from synthetic 3D models to real 3D scenes in this paper (RTL). To solve the model-to-scene and synthetic-to-real issues, we propose a simple yet efficient framework that consists of physical data alignment and convex-hull regularized feature alignment. Extensive experiments on two datasets show that the neural network efficiently recognizes specific categories of objects in a real unseen scene by only learning from synthetic models. Besides, our proposed RTL can also be regarded as a pretext pretraining task that can favour the downstream task's performance.

\clearpage
%
%
\bibliographystyle{splncs04}
\bibliography{egbib}
\end{document}